\documentclass{article}

\usepackage[final,nonatbib]{neurips_2020}

\usepackage[utf8]{inputenc} 
\usepackage[T1]{fontenc}    
\usepackage{hyperref}       
\usepackage{url}            
\usepackage{booktabs}       
\usepackage{amsfonts}       
\usepackage{nicefrac}       
\usepackage{microtype}      
\usepackage{graphicx}
\usepackage{multirow}
\usepackage{chngcntr}

\title{Semantic Segmentation of Medium-Resolution Satellite Imagery using Conditional Generative Adversarial Networks}

\author{%
  Aditya Kulkarni \\
   Radiant Earth Foundation\\
   San Francisco, CA 94105 \\
  \texttt{adkulkar@eng.ucsd.edu} \\
  \And
    Tharun Mohandoss \\
    Radiant Earth Foundation\\
    San Francisco, CA 94105 \\
  \texttt{tharun96@utexas.edu} \\
   \AND
   Daniel Northrup \\
   Benson Hill \\
   Saint Louis, MI \\
   \texttt{dan@northrup.ag} \\
   \And
   \And
   Ernest Mwebaze \\
   Sunbird AI \\
   Kampala, Uganda \\
   \texttt{emwebaze@gmail.com} 
      \And
  Hamed Alemohammad \\
   Radiant Earth Foundation\\
   San Francisco, CA 94105 \\
  \texttt{hamed@radiant.earth} \\
}

\begin{document}

\maketitle

\begin{abstract}

Semantic segmentation of satellite imagery is a common approach to identify patterns and detect changes around the planet. Most of the state-of-the-art semantic segmentation models are trained in a fully supervised way using Convolutional Neural Network (CNN). The generalization property of CNN is poor for satellite imagery because the data can be very diverse in terms of landscape types, image resolutions, and scarcity of labels for different geographies and seasons. Hence, the performance of CNN doesn't translate well to images from unseen regions or seasons. Inspired by Conditional Generative Adversarial Networks (CGAN) based approach of image-to-image translation for high-resolution satellite imagery, we propose a CGAN framework for land cover classification using medium-resolution Sentinel-2 imagery. We find that the CGAN model outperforms the CNN model of similar complexity by a significant margin on an unseen imbalanced test dataset. 
  
\end{abstract}

\section{Introduction}
Semantic segmentation techniques applied to Earth observation (EO) can be helpful in identifying patterns in imagery and detecting changes throughout time. Lately, semantic segmentation algorithms using Convolutional Neural Network (CNN) have produced state-of-the art results on various medium-resolution EO data like Landsat and Sentinel-2~\cite{Knopp_2020, 8519195, 7891032, TONG2020111322}. CNN architectures are designed to capture both the high and the low level details in the images. \cite{arch_cnn_ref, unet_paper} use encoder-decoder type of architecture and skip connections for this. Even though CNN based approaches outperform other hand-engineered feature based solutions, these approaches can lack the generalization required to handle the seasonal and regional diversity of EO imagery~\cite{GeoGAN}. 

 \cite{isola2016imagetoimage} provides an alternative perspective to computer vision problems, by looking at them as an image-to-image translation problem. This is done using Generative Adversarial Networks (GANs)~\cite{goodfellow2014generative} in the conditional setting~\cite{MirzaO14}. GANs are trained to solve a two-player mini-max game, where generator $G$ tries to fool the discriminator $D$ by generating images that resemble the original dataset, while discriminator tries to differentiate the generated images from the original dataset. This results in two architectures trying to beat each other; improve together in the process, until eventually they reach Nash equilibrium or an impasse. The objective function is given by Eq.~\ref{obj_function}, and unlike CNN based approaches does not require problem specific loss functions. Here $x$ is the original dataset sample, $z$ is the noise and $G(z)$ is generated data sample. The discriminator tries to maximize $L_{GAN}$ and the generator tries to minimize $L_{GAN}$.
\begin{equation}
    L_{GAN} =  E_x{[log(D(x))]} + E_z{[log(1-D(G(z))]}
    \label{obj_function}
\end{equation}

Thanks to GAN's promising performance, their applications has emerged in EO too \cite{GeoGAN, hyrda_gan, wgan_for_road}. Further~\cite{GeoGAN} has proposed that this technique can be applied to semantic segmentation problem on satellite imagery, and demonstrated that GANs can achieve near CNN performance. But CNNs still emerge victorious in high resolution imagery like that of~\cite{potsdam}. We find that Conditional Generative Adversarial Networks (CGAN) has the potential to outperform CNN model of similar architectural capacity by a significant margin; when trained on medium-resolution satellite imagery, and tested on unseen imagery from a different location. Further, objects of the same land cover class in satellite imagery (e.g. cropland) are very distinct across different parts of the world. This also requires a segmentation model that can generalize beyond the initial training dataset, as these labeled data are scarce at global scale. 

\section{Dataset}
\begin{figure}
    \centering
    \includegraphics[width=0.9\linewidth]{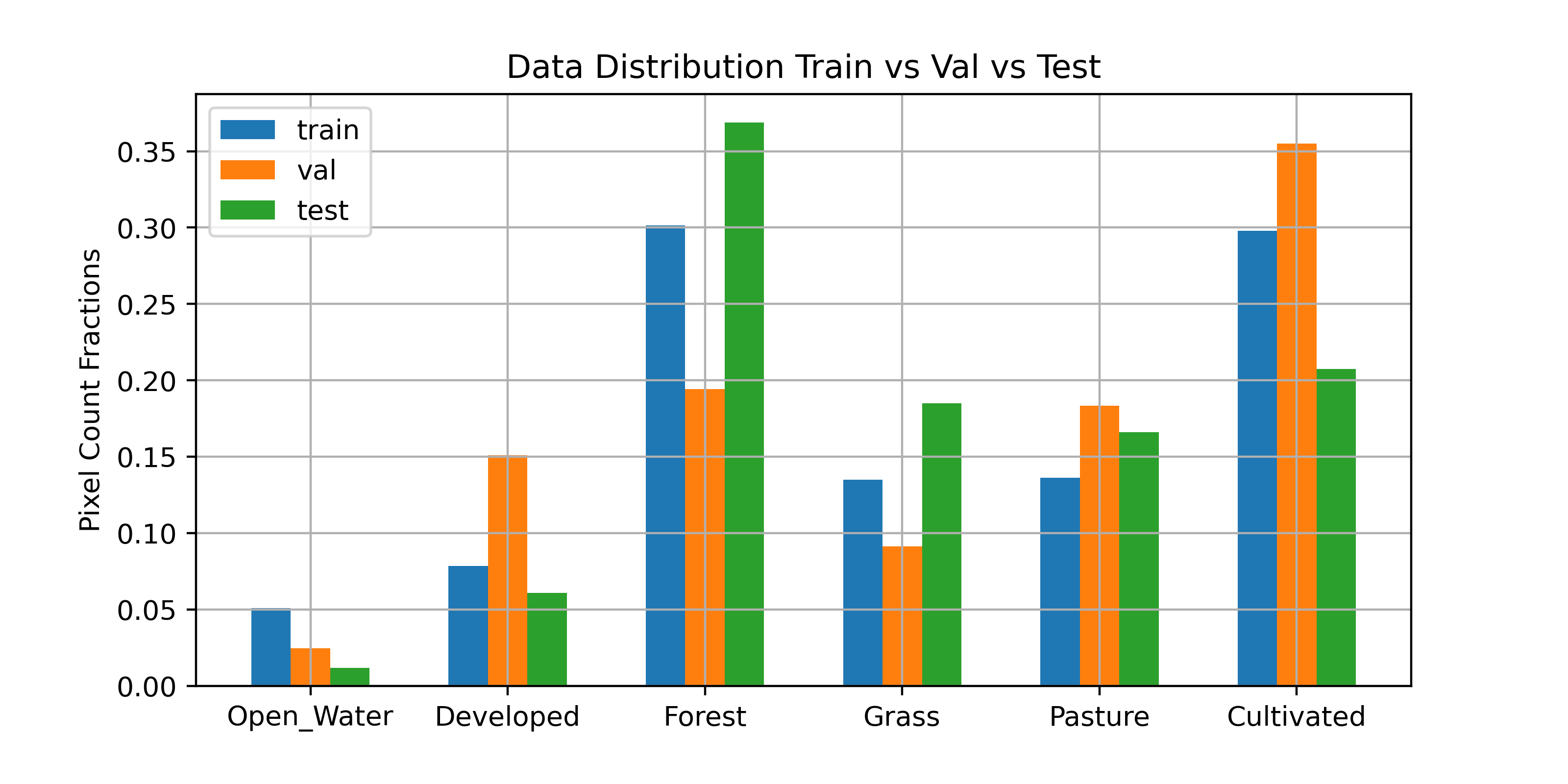}
    \caption{Class distribution in train, validation and test datasets}
    \label{data_trainvstest_distribution}
\end{figure}

\begin{figure}
    \centering
    \includegraphics[width=0.9\linewidth]{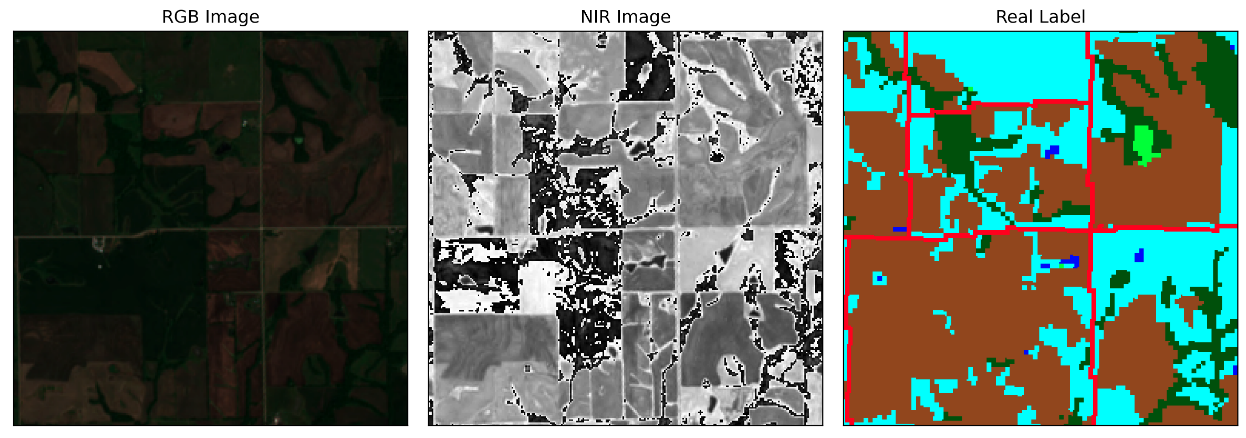}
    \caption{Sample RGB image with corresponding NIR band and labels. (label colors: red: developed, dark blue: open water, cyan: pasture, dark green: forest, light green: grass, brown: cultivated)}
    \label{sample_data}
\end{figure}

In this work, we use Sentinel-2 satellite imagery, which has a resolution of 10 meters. Even though the dataset has a total of 13 spectral bands we choose Red, Green, Blue and Near Infrared (NIR) bands as a first step towards the problem. We use land cover classes from the National Land Cover Database (NLCD) that provides a 16-class semantic labels for each 30m pixel in the Sentinel-2 imagery. We selected multiple regions within the continental US to generate skewed training and test dataset. To simplify the complicated nuances in the similar classes, we apply some post processing on the labels. First, we merge together classes belonging to similar categories: deciduous forest, evergreen forest, mixed forest, and shrubs as forest class. Also a single developed class instead of developed open space, developed low intensity, developed medium intensity, and developed high intensity. Second, we dropped images belonging to extremely rare classes like barren land, woody wetlands and emergent wetlands. Thereby, reducing the total number of classes to six: Open Water, Developed, Forest, Grass, Pasture and Cultivated. Train and validation were selected from same regions, while test datasets was selected from a different region. Details about the locations can be found in the section ~\ref{GeographicalDetails} of the appendix. In total we have around 11943 train, 3989 validation and 6795 test images. The distribution of train, validation and test data is presented Figure~\ref{data_trainvstest_distribution}. A sample image and its corresponding label is shown in Figure~\ref{sample_data}. 

\section{Methods}

Inspired by~\cite{GeoGAN} and~\cite{isola2016imagetoimage}, we use a special variant of GANs, called CGAN, to train the model for segmentation task. The dense image labelling task is looked at as that of an image-to-image translation: from a source image domain $X$ i.e. from RGB+NIR to a target image domain $Y$ i.e. dense labels. Here the generated output $\hat{Y}$ is conditioned on the input $X$. CGANs are popular for their superior performance over CNN in image-to-image translation tasks like image coloration, style transfer and dense labelling~\cite{GeoGAN}. Like GAN, CGAN optimizes a two-person mini-max objective function given by Eq.~\ref{CGAN_loss}. The generator $G$ tries to minimize $L_{CGAN}$ while discriminator $D$ tries to maximize $L_{CGAN}$.

\begin{equation}
    L_{CGAN} = E_{X,Y} (log(D(X,Y))) + E_{X,\hat{Y}}(log(1-D(X,G(X))))
    \label{CGAN_loss}
\end{equation}

Along with $L_{CGAN}$, we use $L_2$ norm loss to train the generator. This helps to capture all the low frequency features from the source images to target images. We found the performance of $L_2$ to be much better than $L_1$ in terms of F1-score, visual perception and also training stability. And since our dataset is imbalanced we use weighted $L_2$ norm loss Eq.~\ref{weighted_L2} where the weight used is inverse of the class fraction in the training set. Hence the overall generator loss is given by Eq.~\ref{overall_loss} where $w_c$ is the fraction of pixels of class $c$, $\lambda = 100$, $Y$ and $\hat{Y}$ are the target labels and generated labels respectively.

\begin{equation}
    L_{L^2}  = \sum_{c=1} ^{6} \frac{\frac{1}{w_c}||Y-\hat{Y}||_2}{\sum_{c=1} ^{6} \frac{1}{w_c}}  
    \label{weighted_L2}
\end{equation}

\begin{equation}
    L_{G} = L_{CGAN} + \lambda L_{L^2}
    \label{overall_loss}
\end{equation}

\subsection{CGAN Model}
Based on the success of benchmarks in pic2pic~\cite{isola2016imagetoimage} and GeoGAN~\cite{GeoGAN}, we use a U-Net type architecture for our generator. We choose a 14 layer deep U-Net for our generator, and use Patch GAN~\cite{patch_gan} with 5 layers deep for our discriminator. Architectural details can be found in the table ~\ref{gen_arch},~\ref{disc_arch} of the appendix. In contrast with GeoGAN, we found that a shallower network was better in learning the difference between classes. This could be because of the lower resolution of Sentinel-2 images compared to \cite{potsdam} used in GeoGAN. We optimize Eq.~\ref{overall_loss} by using Adam optimizer for $G$ with $\beta=(0.5,0.99)$ and plain stochastic gradient descent (SGD) for $D$. Both optimizers use a learning rate of $2e-4$.

\subsection{CNN Model}
We also train a 14-layer deep U-Net in a fully supervised setting using cross entropy loss. This architecture is exactly same as the generator used in CGAN based approach. To have a fair comparison with CGAN architecture, we use a pixel-wise weighted cross entropy loss given by Eq.~\ref{CNN_loss} (in the appendix) to account for class imbalance.

\section{Results and Discussions}
\begin{figure}
    \centering
    \includegraphics[width=0.9\linewidth]{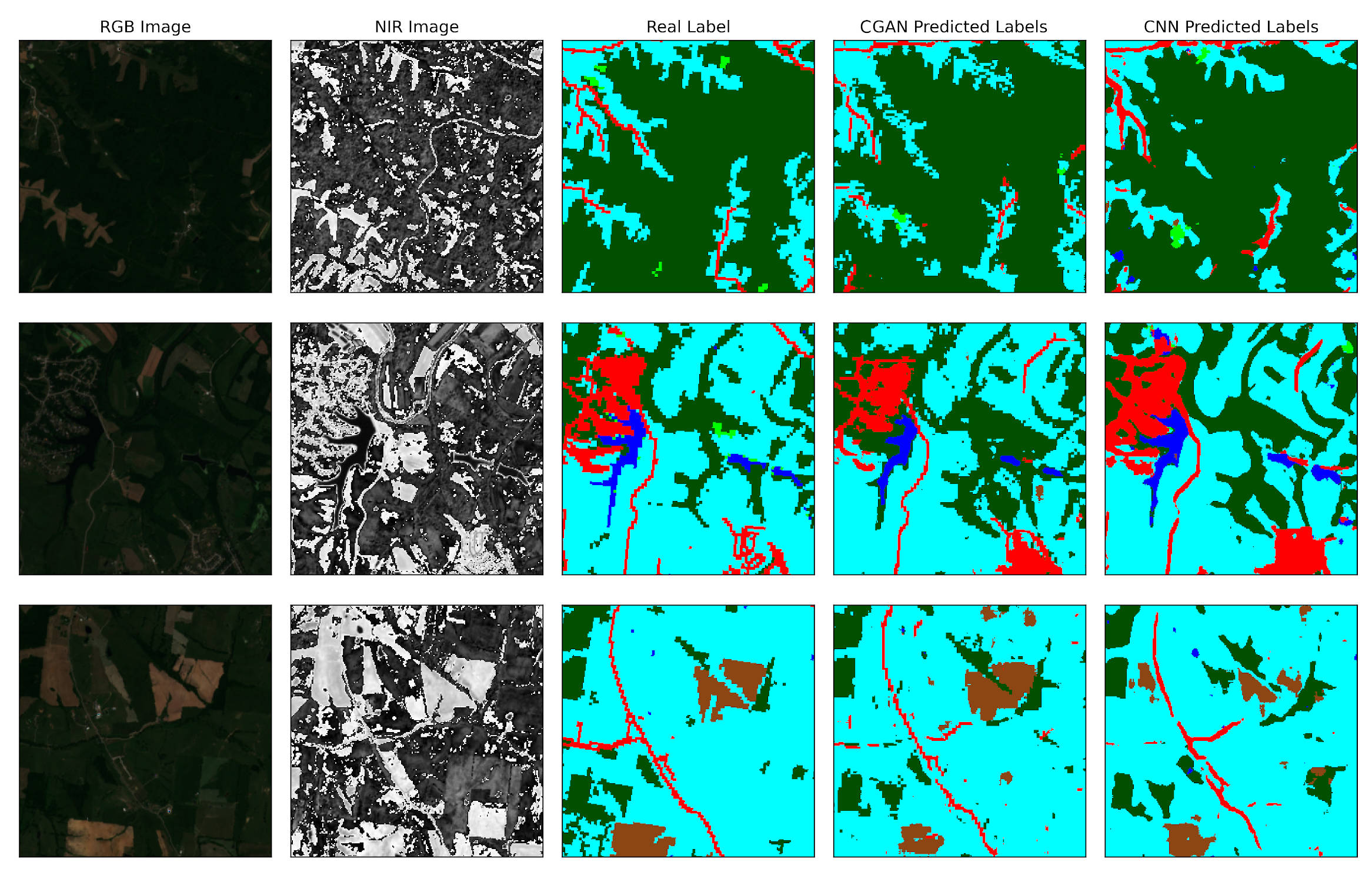}
    \caption{Comparison of CGAN and CNN predictions in the test dataset. (label colors: red: developed, dark blue: open water, cyan: pasture, dark green: forest, light green: grass, brown: cultivated)}
    \label{test_sample_1}
\end{figure}

We split the training dataset into train-fold and validation-fold, and tune for hyper-parameters for CGAN and CNN based architectures. Then the models are tested on an unseen dataset from a totally different geographical location as described in Appendix~\ref{GeographicalDetails}. Table~\ref{f1_scores} presents the F1-score for train, validation and test datasets. We observe that on train and validation there was a tie between the two architectures with each exhibiting dominating performance over 3 classes. However, on test data CGAN outperforms CNN in five out of six classes. Figure~\ref{test_sample_1} shows 3 examples of the input imagery, true labels, and predicted labels from CGAN and CNN. More examples is provided in Figure~\ref{extra_test_gan_vs_cnn}

\begin{table}
   \caption{F1 Scores for train, validation and test datasets}
    \label{f1_scores}
    \centering
    \begin{tabular}{cccccccc} 
     \toprule
     Set & Architecture & Open Water & Developed & Forest & Grass & Pasture & Cultivated\\ 
     \midrule
     \multirow{2}{*}{Train} & CGAN  & \textbf{97.686}  & 76.371  & 78.939 &  \textbf{68.553} & \textbf{65.518} &  84.569\\ 
     & CNN & 91.578  &   \textbf{84.899}  &  \textbf{81.443} & 62.734  & 57.784 & \textbf{90.693}\\
     \midrule
     \multirow{2}{*}{Validation} & CGAN & \textbf{94.872}  &   83.537 &70.247  &  \textbf{39.265} & \textbf{49.682}  & 81.899 \\ 
     & CNN  & 84.291  & \textbf{84.029}  &  \textbf{71.858} &  36.511& 32.258 &  \textbf{85.956}\\
     \midrule
     \multirow{2}{*}{Test} & CGAN & \textbf{82.985}  &   \textbf{54.402} &\textbf{62.093}  &  34.382 & \textbf{56.655}  & \textbf{59.197} \\ 
     & CNN  & 49.081  &50.343  &  61.635&  \textbf{40.816}& 40.203 &  57.606\\ 
    \bottomrule
    \end{tabular}
\end{table}

The performance enhancement observed in CGAN based approach demonstrated better generalization capability of the adversarial training over fully supervised training - for similar architectures. The CGAN based approach uses a total of 44.6M parameters, while CNN based approach uses 41.83M parameters. Hence, we can see that by using an additional 6.62\% parameters we can achieve significant performance improvement on unseen test data. Additionally, we can observe from the predicted labels that the CGAN generated labels have learned all the nuances present in the target labels. The contours of objects in the CGAN labels are much similar to the target labels than the CNN based approach even on test data (Figure~\ref{test_sample_1}). Here, we demonstrated the performance gain of CGAN over CNN on unseen data and, as a future work we can explore the performance benefits of CGAN over CNN in a wider transfer learning setting: from a label rich region like the US to label scarce regions like Africa and Asia. We believe this approach can also be a good step towards building semantic models to produce labels across different seasons as well.

\section*{Acknowledgements}
This research is funded by a grant awarded to Radiant Earth Foundation through the 2019 Grand Challenges Annual Meeting Call-to-Action from the Bill \& Melinda Gates Foundation. The findings and conclusions contained within are those of the authors and do not necessarily reflect positions or policies of the Bill \& Melinda Gates Foundation.

\bibliographystyle{unsrt}
\bibliography{neurips_2020}

\newpage
\appendix
\counterwithin{figure}{section}
\counterwithin{table}{section}
\counterwithin{equation}{section}

\section{Network Architectures}
Our experiments are performed using PyTorch libraries and relies on the code released by pix2pix \cite{pix2pix2017}. The architectural details that we considered for the CGAN and CNN are listed in table \ref{gen_arch}, \ref{disc_arch}. The generator architecture is also the same one used for CNN. Here, we use 4 $\times$ 4 kernel for performing both 2D convolutions (Conv2D) and transpose 2D convolutions. LReLU stands for leaky relu with negative slope = $0.2$, BN stands for batch normalization and DO stands for drop out with probability $0.5$.

\begin{table}[h]
   \caption{Generator Architecture}
    \label{gen_arch}
    \centering
    \begin{tabular}{cccc} 
     \toprule
     Block & Input Shape & Operations & Output Shape\\ 
     \midrule
     1. &  4 x 256 x 256      & Conv2D, LReLU &   64 x 128 x 128 \\ 
     2. & 64 x 128 x 128      & Conv2D, BN, LReLU & 128 x 64 x 64  \\
     3. & 128 x 64 x 64      & Conv2D, BN, LReLU & 256 x 32 x 32  \\
     4. & 256 x 32 x 32      & Conv2D,  BN, LReLU & 512 x 16 x 16  \\ 
     5. & 512 x 16 x 16      & Conv2D,  BN, LReLU & 512 x 8 x 8  \\ 
     6. & 512 x 8 x 8      & Conv2D,  BN, LReLU & 512 x 4 x 4  \\ 
     7.& 512 x 4 x 4      & Conv2D, ReLU & 512 x 2 x 2  \\ 
     8.& 512 x 2 x 2      & ConvTrans2D,  BN, ReLU & 512 x 4 x 4  \\ 
     9.& 1024 x 4 x 4      & ConvTrans2D,  BN, DO, ReLU & 512 x 8 x 8  \\ 
     10.& 1024 x 8 x 8      & ConvTrans2D,  BN,DO, ReLU & 512 x 16 x 16  \\
     11.& 1024 x 16 x 16      & ConvTrans2D,  BN, ReLU & 256 x 32 x 32  \\
     12.& 512 x 32 x 32      & ConvTrans2D,  BN, ReLU & 128 x 64 x 64  \\
     13.& 256 x 64 x 64      & ConvTrans2D,  BN, ReLU & 64 x 128 x 128  \\
     14.& 128 x 128 x 128      & ConvTrans2D,  Softmax & 6 x 256 x 256  \\
    \bottomrule
    \end{tabular}
\end{table}

\begin{table}[h]
   \caption{Discriminator Architecture}
    \label{disc_arch}
    \centering
    \begin{tabular}{cccc} 
     \toprule
     Block & Input Shape & Operations & Output Shape\\ 
     \midrule
     1. &  10 x 256 x 256      & Conv2D, LReLU &   64 x 128 x 128 \\ 
     2. & 64 x 128 x 128      & Conv2D, BN, LReLU & 128 x 64 x 64  \\
     3. & 128 x 64 x 64      & Conv2D, BN, LReLU & 256 x 32 x 32  \\
     4. & 256 x 32 x 32      & Conv2D,  BN, LReLU & 512 x 16 x 16  \\ 
     5. & 512 x 16 x 16      & Conv2D,  Sigmoid & 1 x 8 x 8  \\ 
    \bottomrule
    \end{tabular}
\end{table}

\section{CNN Optimization}

We optimize a weighted cross entropy loss as in Eq. ~\ref{CNN_loss} where $w_c$ is the fraction of pixels of class $c$  in the train dataset, and $\hat{y}_{pred}$ and $y_{true}$ are the predicted class probability and one-hot encoded target respectively.  We optimize the  Eq.~\ref{CNN_loss} by using Adam optimizer with $\beta=(0.5,0.99)$ and a learning rate of $2e-4$.

\begin{equation}
    L_{CNN} = -E_{pixels}[\sum_{c=1} ^6 \frac{y_{true}}{w_c}  log(\hat{y}_{pred})]
    \label{CNN_loss}
\end{equation}

\section{Geographical Details of the Training and Test Datasets}
\label{GeographicalDetails}

In order to build the training dataset, we ensured that there were non-zero number of pixel samples for each of the six classes. But, as the dataset collected from a single city or region always tends to have highly skewed class distribution, we collected both training and test data from a number of regions around the US. To build the training dataset we sampled images from Iowa, Sacramento, Dallas, Chicago, Los Angeles, San Diego, Houston, New York, Boston, Baltimore, Philadelphia, Seattle and Detroit areas. Similarly, to build the test dataset we considered Montana, Kentucky, Green Bay Wisconsin and area around Lake Michigan.  

\section{Additional Test Image}

\begin{figure}
    \centering
    \includegraphics[width=0.9\linewidth]{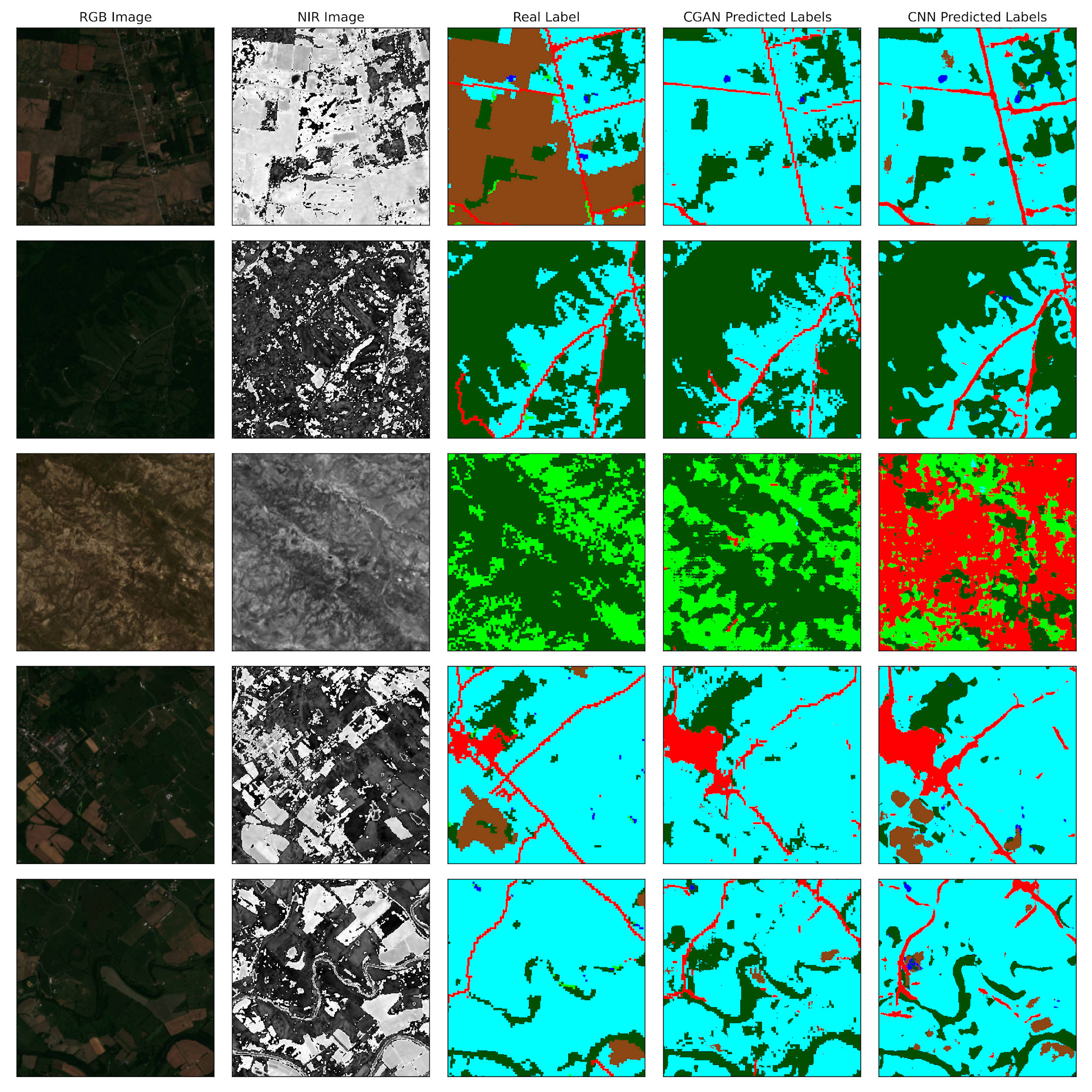}
    \caption{More examples similar to Figure~\ref{test_sample_1}. (label colors: red: developed, dark blue: open water, cyan: pasture, dark green: forest, light green: grass, brown: cultivated)}
    \label{extra_test_gan_vs_cnn}
\end{figure}

More predicted classes from the test dataset are shown in Figure~\ref{extra_test_gan_vs_cnn}. We can see that the models struggle with cultivated class in the test set. It is worth noting that the CGAN model performs better on grass class in the test data (different region), while CNN confuses such patterns with development class. Even visually CGAN generated labels are more similar to the real labels than the CNN generated labels. 

\end{document}